\documentclass{article} 
\usepackage{iclr2024_conference,times}
\usepackage{graphicx,subcaption,amsmath,amssymb,array,balance,hyperref,multicol,multirow,siunitx,times,subcaption,mathtools,colortbl,wrapfig}

\usepackage{amsmath,amsfonts,bm}

\def\eqref#1{equation~\ref{#1}}

\def\1{\bm{1}}

\DeclareMathAlphabet{\mathsfit}{\encodingdefault}{\sfdefault}{m}{sl}
\SetMathAlphabet{\mathsfit}{bold}{\encodingdefault}{\sfdefault}{bx}{n}

\newcommand{\E}{\mathbb{E}}

\DeclareMathOperator*{\argmin}{arg\,min}

\newcommand{\cmt}[1]{}

\newcommand{\motion}{\mathbf{m}}

\newcommand{\sr}{\mathbf{s}^r}
\newcommand{\xh}{\mathbf{x}^h}

\newcommand{\ph}{\mathbf{p}^h}
\newcommand{\pr}{\mathbf{p}^r}

\long\def\ignorethis#1{}

\newcommand{\pctab}{\hspace{0.2in}}

\usepackage{hyperref}
\usepackage{url}
\usepackage{algorithm}
\usepackage{tabularx}

\usepackage[noend]{algpseudocode}
\usepackage[export]{adjustbox}

\title{CrossLoco: Human Motion Driven Control of Legged Robots via Guided Unsupervised Reinforcement Learning}

\author{Tianyu Li, Hyunyoung Jung, Matthew Gombolay, Yong Kwon Cho, Sehoon Ha  \\
Georgia Institute of Technology\\
Atlanta, GA 30332, USA \\
\texttt{\{tli471,hjung331,sehoonha\}@gatech.edu} \\
\texttt{matthew.gombolay@cc.gatech.edu} \\
\texttt{yong.cho@ce.gatech.edu} \\
}

\iclrfinalcopy 
\begin{document}

\maketitle

\begin{abstract}
Human motion driven control (HMDC) is an effective approach for generating natural and compelling robot motions while preserving high-level semantics. However, establishing the correspondence between humans and robots with different body structures is not straightforward due to the mismatches in kinematics and dynamics properties, which causes intrinsic ambiguity to the problem. Many previous algorithms approach this motion retargeting problem with unsupervised learning, which requires the prerequisite skill sets. However, it will be extremely costly to learn all the skills without understanding the given human motions, particularly for high-dimensional robots. In this work, we introduce \textit{CrossLoco}, a guided unsupervised reinforcement learning framework that simultaneously learns robot skills and their correspondence to human motions. Our key innovation is to introduce a cycle-consistency-based reward term designed to maximize the mutual information between human motions and robot states. We demonstrate that the proposed framework can generate compelling robot motions by translating diverse human motions, such as running, hopping, and dancing. We quantitatively compare our \textit{CrossLoco}  against the manually engineered and unsupervised baseline algorithms along with the ablated versions of our framework and demonstrate that our method translates human motions with better accuracy, diversity, and user preference. We also showcase its utility in other applications, such as synthesizing robot movements from language input and enabling interactive robot control.
\end{abstract}

\section{Introduction}

The concept of teleoperating robots through human movements,
known as Human Motion Driven Control (HMDC), has been illustrated in various forms of media, including animations, movies, and science fiction, such as Madö King Granzört~\citep{Madö_King_Granzört}, Pacific Rim~\citep{Pacific_Rim}, and Ready Player One~\citep{Ready_Player_One}. In these media, HMDC technology allows operators to intuitively control robots using their body movements. Compared to fully autonomous control, this teleoperation offers the essential dexterity and decision-making capabilities required for tasks demanding precise motor skills and situational awareness. Consequently, this property makes HMDC promising for various applications, including entertainment, medical surgery, and space exploration.

The key challenge of HMDC is how to establish the correspondence between robot states and human motions, which can also be referred to as motion retargeting. For certain types of robots, such as humanoids or manipulators, this correspondence might be simple enough to be approached by assuming the mapping of end-effectors in Cartesian space and solving the formulated inverse kinematics problem~\citep{gleicher1998retargetting, tak2005physically}. However, when we consider robots with significantly different morphological structures, such as quadrupeds, hexapods, or quadrupeds with mounted arms, the correspondence becomes nontrivial due to the intrinsic ambiguity of the problem. Therefore, researchers often have approached this motion retargeting problem by applying supervised learning techniques to the paired datasets~\citep{sermanet2018time,delhaisse2017transfer,rhodin2014interactive}. Nonetheless, creating paired datasets can be a challenging and labor-intensive task that requires significant engineering expertise. To address this issue, some researchers have proposed using unsupervised learning techniques to learn the correlation from unpaired human and robot motion datasets~\citep{li2023ace, choi2020nonparametric, smith2019avid}. In this case, the robot dataset serves as prior knowledge indicating the motion pattern of the robot. However, obtaining motion datasets can be expensive because we do not know the required skills for the given human motion. In addition, control itself is challenging due to the complexity of the quadrupedal robot and its underactuated dynamics. This leads to our research question: can we learn cross-morphology HMDC without prior knowledge of the robot?

The research question presents three primary challenges. Firstly, the significant difference in kinematics and dynamics between the human and the target robot makes it difficult to establish correspondence. Secondly, we cannot build a predefined motion database for the robot due to the complexity of the problem. Finally, the problem itself is ambiguous. For instance, there exist many different quadrupedal gaits that can capture the essence of human walking. To address these challenges, we drew inspiration from the recent unsupervised skill discovery techniques, such as~\cite{eysenbach2018diversity} and~\cite{2022-TOG-ASE}, and aim to simultaneously learn robot skills and robot-human motion correspondence by maximizing the mutual information between human and robot motions.

\begin{figure*}
\vspace{-1em}
\centering
\includegraphics[width=.995\linewidth]{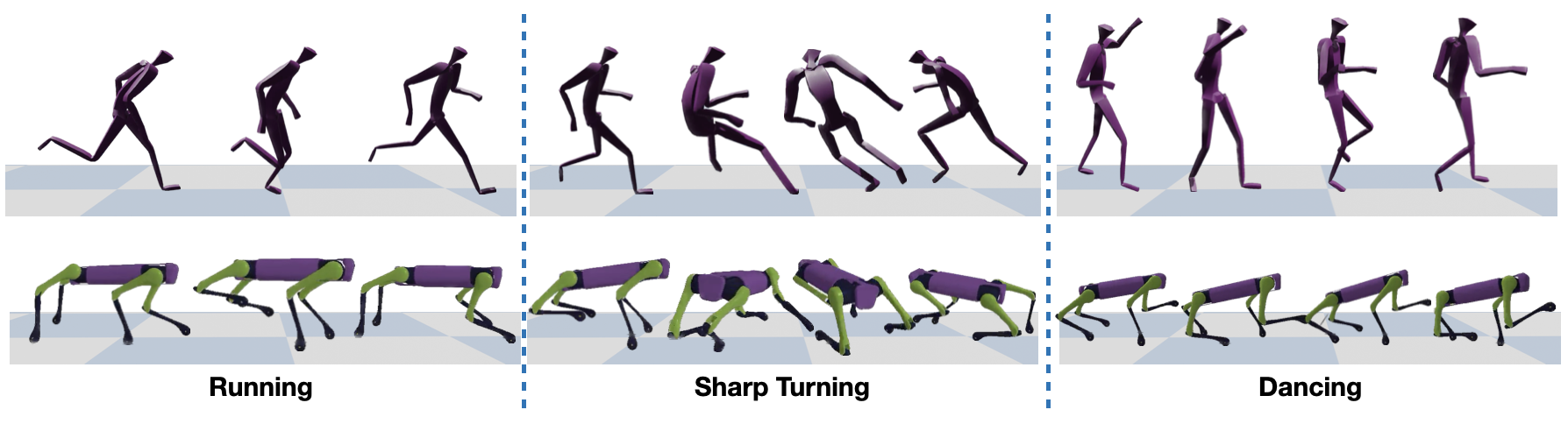}
\caption{We introduce \textit{CrossLoco}, a guided unsupervised reinforcement learning framework for translating human motion to robot control.
}
\label{fig:overview_demo}
\end{figure*}

In this work, we introduce \textit{CrossLoco}, a guided unsupervised reinforcement learning framework that enables simultaneous learning of human-robot motion correspondence and robot motion control (Figure~\ref{fig:overview_demo}). Our key approach is to introduce a cycle-consistency-based correspondence reward term that maximizes the mutual information between human motions and the synthesized robot movements. We implement this cycle consistency term by training both robot-to-human and human-to-robot reconstruction networks. Our formulation also includes regularization terms and a root-tracking reward to guide correspondence learning. Simultaneously, we train a robot control policy that takes human motions and sensory information as input and generates robot actions for interacting with the environment. 

We demonstrate that \textit{CrossLoco} can translate a large set of human motions for robots, including walking, running, and dancing. Even for locomotion, the robot exhibits two distinct strategies, trotting and galloping, inspired by human walking motions with different styles. We quantitatively compare our method against the baseline, DeepMimic~\citep{2018-TOG-deepMimic}, along with the ablated versions of our \textit{CrossLoco} framework and show that our method can achieve better quantitative results in terms of accuracy, diversity, and user preference. We further showcase the potential applications of our framework: \textbf{language2text} motion synthesis and \textbf{interactive motion control}.

\section{Related Works}

\textbf{Learning Locomotion Skills}. There are various methods for robots to learn locomotion skills. One approach involves maximizing a reward function designed by experts using reinforcement learning, as demonstrated in several studies such as~\cite{tan2018sim, haarnoja2018learning, xie2018feedback, li2019using, rudin2022learning}. Another method is motion imitation, where the control policy is trained with an imitation reward, as shown in~\cite{2018-TOG-deepMimic,2018-TOG-SFV,li2023fastmimic,bergamin2019drecon,won2019learning,ling2020character,RoboImitationPeng20}. This reward is calculated based on the distance between the robot's current pose and a reference pose from the demonstration trajectory. The closer the distance, the larger the reward. Generative adversarial imitation learning (GAIL)~\citep{ho2016generative} is another approach that trains the policy to deceive a discriminator that distinguishes real and fake demonstration data. Finally, without the need for engineering reward functions or demonstration data, some studies such as~\cite{eysenbach2018diversity, sharma2019third} focus on unsupervised skill discovery from interaction data through information-theoretic methods.

\textbf{Motion Retargeting}.
Transferring motions between different morphologies has been an important topic in both robotics and computer graphics communities to produce natural motions for various robots and characters. Researchers have investigated various approaches, such as designing manual correspondences~\citep{gleicher1998retargetting,tak2005physically,grandia2023doc}, learning from paired datasets~\citep{sermanet2018time,delhaisse2017transfer,jang2018variational}, or developing modular/hierarchical policies~\citep{won2019learning,hejna2020hierarchically,sharma2019third}. More recent works~\citep{zhang2020learning,aberman2020skeleton,villegas2018neural,li2023ace,smith2019avid,kim2020domain,shankar2022translating} aim to learn the state and action correspondence from unpaired datasets via unsupervised learning. However, these methods often require a pre-collected dataset of both domains, which is not available for robots in our problem.
\textbf{Cycle-Consistency}. Our work is inspired by previous research on cycle-consistency~\citep{zhou2016learning, zhu2017unpaired,liu2017unsupervised,rao2020rl,bousmalis2018using}. For instance, CycleGAN~\citep{zhu2017unpaired} combines cycle-consistency loss with Generative Adversarial Networks~\citep{goodfellow2014generative} for unpaired image-to-image translation. By adding domain knowledge, CycleGAN can be extended to video retargeting~\citep{bansal2018recycle} and domain adaptation~\citep{hoffman2018cycada}. In robotics, a similar approach has been investigated for sim-to-real transfer~\citep{stein2018genesis,james2019sim}. Besides alignment in image space, a few researchers~\citep{zhang2020learning,shankar2022translating} 
adopt cycle-consistency to align agents in different dynamics and structures, while the others~\citep{aberman2020skeleton,villegas2018neural} apply cycle-consistency for motion retargeting between similar human-like robots or characters. Inspired by these works, we aim to co-train a control policy for the diverse motor skills of a quadrupedal robot while establishing cycle consistency between the robot and human motions.

\section{Preliminaries}
\textbf{Skill-Conditioned Reinforcement Learning.}
We formulate our framework as a skill-conditioned reinforcement learning problem, where an agent interacts with an environment to maximize an objective function by following a policy $\pi$. At the beginning of each learning episode, a condition term is sampled from the dataset $\mathbf{z} \sim p(\mathbf{z})$. At each time step, the agent observes the state of the system $\mathbf{s}_t$, then takes an action sampled from the policy  $\mathbf{a}_t \sim \pi(\mathbf{a}_t|\mathbf{s}_t, \mathbf{z})$ to interacts with the environment. After executing the actions, the environment takes the agent to a new state sampled from the dynamics transition probability $\mathbf{s}_{t+1} \sim p(\mathbf{s}_{t+1}|\mathbf{s}_{t}, \mathbf{a}_t)$. A scalar reward can be measured using a reward function $r_t = r(\mathbf{s}_{t}, \mathbf{a}_t, \mathbf{s}_{t+1}, \mathbf{z})$. The agent's objective is to learn a policy that maximizes its expected cumulative reward $J(\pi)$,
\begin{align}
\label{equ:GoalConditionedRL}
    J(\pi) = \E_{\mathbf{z} \sim p(\mathbf{z}), \bm{\tau} \sim p(\bm{\tau}|\pi, \mathbf{z})}[\sum^{T-1}_{t=0}\gamma^t r_t].
\end{align}
Here, $\bm{\tau}$ is a state and action trajectory with the length $T$, where its distribution can be computed as $p(\bm{\tau}|\pi, \mathbf{z}) = p(\mathbf{s}_0) \prod^{T-1}_{t=0}p(\mathbf{s}_{t+1}|\mathbf{s}_t, \mathbf{a}_t)\pi(\mathbf{a}_t|\mathbf{s}_t, \mathbf{z})$ is the likelihood of the trajectory under policy $\pi$. The initial state $\mathbf{s}_0$ is sampled from the distribution $p(\mathbf{s}_0)$ and $\gamma \in [0,1)$ is a discount factor.

\textbf{Skill Discovery By Maximizing Mutual Information.}
\cite{eysenbach2018diversity} and~\cite{2022-TOG-ASE} formulate the skill discovery problem as an unsupervised reinforcement learning problem, where the objective is to maximize the mutual information between the robot state and a latent vector sampled from a distribution $\mathbf{z} \sim p(\mathbf{z} )$: $I(\mathbf{S}; \mathbf{Z}) = H(\mathbf{S}) - H(\mathbf{S}, \mathbf{Z})$.
This equation can be interpreted as the policy $\pi$ is to learn to produce diverse behaviors while each latent vector $\mathbf{z}$ should correspond to distinct robot states. 

However, this equation is intractable in most scenarios where the state marginal distribution is unknown, and two tricks are commonly implemented to tackle this. 
The first trick is to take advantage of the symmetry of mutual information:
\begin{align}
\label{equ:sys-MI}
    I(\mathbf{S}; \mathbf{Z}) = I(\mathbf{Z}; \mathbf{S}) = H(\mathbf{Z})-H(\mathbf{Z}| \mathbf{S}).
\end{align}
This trick removes the need for measuring the marginal entropy of robot state $H(\mathbf{S})$ by instead measuring the entropy of the latent vector $H(\mathbf{Z})$ which remains constant in fixed skill prior $p(\mathbf{z})$.    
The second trick is to use a variational lower bound as proposed by ~\cite{eysenbach2018diversity} and ~\cite{gregor2016variational} to approximate the mutual information as follows:
\begin{align}
    \label{equ:lower_bound}
    I(\mathbf{Z}; \mathbf{S}) =  H(\mathbf{Z}) - H(\mathbf{Z}|\mathbf{S}) \ge \max_q H(\mathbf{z} ) + \E_{\mathbf{z}\sim p(\mathbf{z}), \mathbf{s}\sim p(\mathbf{s}|\pi_z)}[\log(q(\mathbf{z}|\mathbf{s})],
\end{align}
where $q(\mathbf{z}|\mathbf{s})$ is a variational approximation  of the conditional distribution $p(\mathbf{z}|\mathbf{s})$ and the lower bound is tight if  $q=p$.
This skill discovery objective encourages a policy to produce distinct behaviors for different skill vectors $\mathbf{z}$ by designing a reward based on the measurement of $q(\mathbf{z}|\mathbf{s})$.
\section{CrossLoco}

\begin{figure*}
\centering
\includegraphics[width=.995\linewidth]{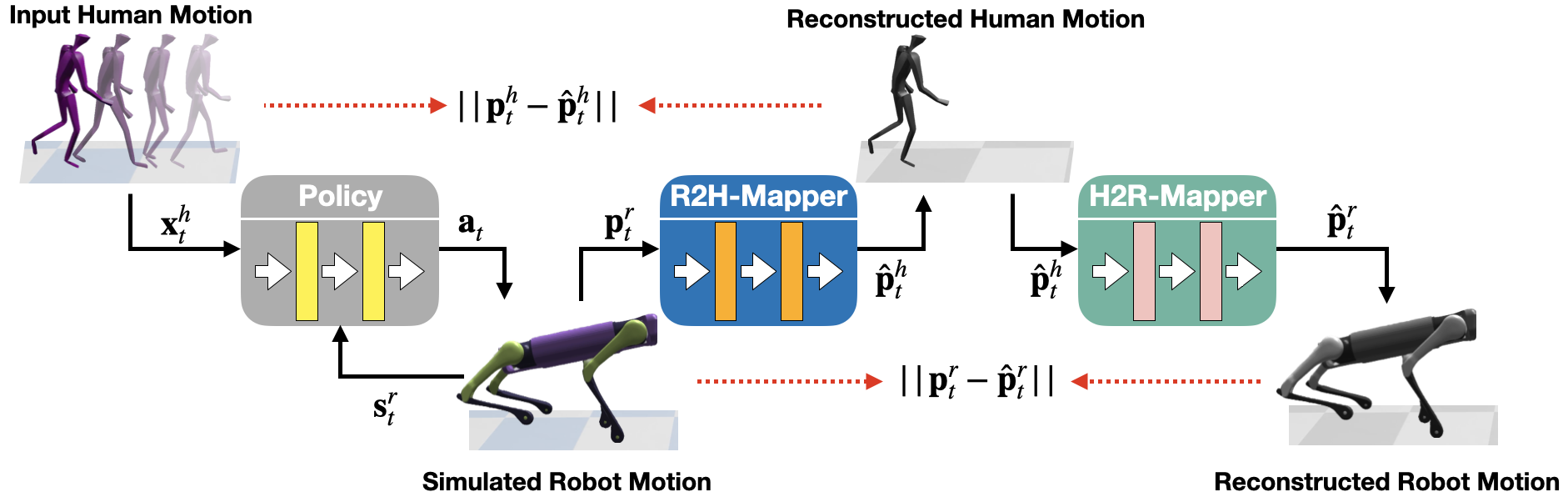}
\caption{Method Overview. A robot control policy utilizes human motions and sensory information to generate robot actions for interacting with the environment. The Robot2Human-Mapper is then used to reconstruct the input human motion pose from the robot's pose. Lastly, the Human2Robot-Mapper is used to reconstruct the robot pose from the reconstructed human pose. Both mappers are trained via supervised learning using the difference between the real and reconstructed poses. These differences are also utilized for constructing a correspondence reward function, which is used to update the policy (see Equation \ref{eqn: crosspondence_reward}).
}
\label{fig:method_overview}
\end{figure*}

\textit{CrossLoco} is a guided unsupervised reinforcement learning framework designed to learn robot locomotion control policy driven by human motion. The framework establishes a correspondence between human and robot motions, enabling the robot to acquire locomotion skills from human motions. The method overview is shown in Fig \ref{fig:method_overview}. In this section, we first introduce how we formulate the problem. Then, we present our cycle-consistency-based method for learning locomotion and human-robot correspondence. Lastly, we provide additional implementation details.

\subsection{Problem Formulation}
Our goal is to train a robot control policy, denoted as $\pi$, that can produce various robot motions based on different human motion inputs. Therefore, we can view our problem as a Markov Decision Process conditioned on the given human motion. Let us define $\ph_t$ and $\pr_t$ as the human and robot kinematic poses. Then the human motion is defined as a sequence of pose vectors: $\motion^h = [\ph_0, \ph_1, \cdots, \ph_{T-1}]$. The robot state $\sr_t$ represents both the kinematic and dynamic status of the robot, hence we can view $\sr_t$ as the superset of the pose $\pr_t$. The robot action $\mathbf{a}_t$ corresponds to motor commands, such as target joint angles. At each time step, the policy takes the robot state vector $\sr_t$ and the augmented human motion feature $\xh_t = x(\ph_t)$ as input to generate an action $\mathbf{a}_t \sim \pi(\mathbf{a}_t | \sr_t, \xh_t)$. Then our goal is to maximize the given reward function $r$:
\begin{align}
\label{equ:real_J}
    J(\pi) = \E_{\motion^h \sim p(\motion^h),\bm{\tau^r} \sim p(\bm{\tau^r}|\pi, \xh_t)}[\sum^{T-1}_{t=0}\gamma^t r(\sr_t, \xh_t,  \mathbf{a}_t)],
\end{align}
where the robot trajectory is defined as a sequence of the robot states $\bm{\tau}^r = [\sr_0, \sr_1, \cdots, \sr_{T-1}]$.

This formulation leads to the question of designing an effective reward function $r$ that builds the relationship between the human and robot poses, $\ph_t$ and $\pr_t$. In addition, the reward function should include some regularization terms, such as minimizing the energy, avoiding self-collisions, or preserving the predefined semantic features. We will discuss the design of our reward function in the following section.

\subsection{Measuring Correspondence via Cycle-consistency}

To develop a reward function that represents the correspondence between human and robot motions, we borrow the information-theoretic approach mentioned in the previous section. 
We formulate the correspondence reward term such that it maximizes the mutual information between human and robot pose, given by $I(\pr_{t}, \ph_t |\pi)$.
From Equation~\ref{equ:lower_bound}, this formulation can be approximated by:
\begin{align}
   I(\pr_{t}, \ph_t |\pi) \geq H(\mathbf{p}^h_t ) + \E_{\motion^h \sim p(\motion^h), \mathbf{p}^r_t\sim p(\mathbf{p}^r_t|\ph_t,\pi)}[\log(q^{r2h}(\mathbf{p}^h_t|\mathbf{p}^r_{t}))]],
\end{align}
where we assume that human motion prior is from the fixed dataset and refer to $q^{r2h}$ as Robot-to-Human Mapper (\textit{R2H-Mapper}). 
Because the first term $H(\mathbf{p}^h_t )$ is constant, we can find the optimal policy by maximizing the second term, $\log[q^{r2h}(\ph_t|\pr_{t})]$.
A higher value represents that \textit{R2H-Mapper} is more certain about the human pose given the robot pose, hence indicating that the human pose is distinctive given the robot pose.

We model the \textit{R2H-Mapper} as a Gaussian distribution with fixed covariance $q^{r2h}(\ph_t|\pr_{t}) = N(\mu^{r2h}(\pr_{t}), \sigma)$ where $\mu^{r2h}(\pr_{t})$ is the mean of the distribution while $\sigma$ is the constant covariance matrix. The \textit{R2H-Mapper} can be trained by minimizing a loss function $ L^{r2h}$:
\begin{align}
   \argmin_{q^{r2h}}\ L^{r2h} = \E_{\ph_t \sim p(\ph_t), \pr_{t} \sim d^\pi(\pr_{t}|\ph_t)}[||\ph_t - \mu^{r2h}(\pr_{t})  ||^2_2],
\end{align}
where $d^\pi(\pr_{t}|\ph_t)$ is the likelihood of observing robot pose $\pr_{t}$, by executing policy $\pi$ given the human pose $\ph_t$. 
Similarly, we can design our correspondence reward to minimize the given term $||\ph_t - \mu^{r2h}(\pr_{t})  ||^2_2$.

However, \textit{R2H-Mapper} does not prevent multiple robot poses $\pr$ from being mapped to the same human pose $\ph$ because it only considers one-directional mapping, which may cause degenerated motions. To address this issue,  we add a Human-to-Robot Mapper (\textit{H2R-Mapper}), denoted as $q^{h2r}(\pr_{t}|\ph_t)$, which is used for mapping the human pose back to the robot pose. We use a cycle-consistency formulation of \citet{zhu2017unpaired}, where we first map the robot pose to the human pose, followed by mapping the generated human pose back to the robot pose. This results in an objective loss function $ L^{r2r}$ for \textit{H2R-Mapper} and \textit{R2H-Mapper} as:
\begin{align}
    \label{eqn: mapper_loss}
   \argmin_{q^{r2h}, q^{h2r}}\ L^{r2r} = L^{r2h} + \E_{\ph_t \sim p(\ph_t), \pr_{t} \sim d^\pi(\pr_{t}|\ph_t)}[ ||\pr_{t} - \mu^{h2r}(\mu^{r2h}(\pr_{t}) ) ||^2_2].
\end{align}

Finally, from our mutual information maximization and cycle-consistency loss minimization, we formulate the correspondence reward as follows:
\begin{align}
\label{eqn: crosspondence_reward}
   r^{cpd}_t = \exp( - ||\ph_{t} -\mu^{r2h}(\pr_{t})  ||^2_2 - ||\pr_{t} - \mu^{h2r}(\mu^{r2h}(\pr_{t}) ) ||^2_2).
\end{align}

\subsection{Implementation Details}

\begin{algorithm}[t]
  	\caption{\footnotesize CrossLoco pseudocode}
  	\label{alg:CrossLoco}
  	\begin{algorithmic}[1]{
\footnotesize
          \Require Human Dataset $\mathcal{M}$.
  	 \State Initialize: Policy $\pi$, Value function $V$, \textit{R2H-Mapper}  $q^{r2h}$, \textit{H2R-Mapper}  $q^{h2r}$, Data Buffer $\mathcal{D}$.

    \Repeat
     \For {trajectory i = 1, ..., m }{\\
     \qquad \quad $\mathcal{D} \leftarrow \{(x^h_t, s^r_t, a_t, r_t)^{T-1}_{t=0}\}$ collect trajectory by rolling out $\pi$. 
    }
    \EndFor
    \State update $\pi$ and $V$ using PPO with data from $\mathcal{D}$.
    \State update $q^{r2h}$ and $q^{h2r}$ using data from $\mathcal{D}$ by minimizing $L^{r2r}$ (Equation~\ref{eqn: mapper_loss}).
    \State Reinitialize Data Buffer $\mathcal{D}$.
\Until{Done}
}
  	\end{algorithmic}
\end{algorithm}


During the training process, the policy $\pi$, as well as the H2R-Mapper and R2H-Mapper, are updated iteratively. The policy is trained using Proximal Policy Optimization (PPO) \cite{schulman2017proximal}. Meanwhile, the H2R-Mapper and R2H-Mapper are trained using supervised learning. The learning framework is summarized in Algorithm \ref{alg:CrossLoco}. 

\textbf{Model Representation.} Human pose ($\ph \in R^{23}$) and robot pose ($\pr \in R^{17}$) consist of local information, including root height, root orientation, and joint pose. The robot state ($\sr_t \in R^{47}$) contains all the information in $\pr$, as well as root and joint velocity, and previous action. The human feature vector ($\xh_t \in R^{188}$) includes human pose, root velocity, and joint velocity information at the future 1, 2, 10, and 30 frames.

\label{sec: implementation_detail}
\textbf{Complete Reward Function.} In addition to the correspondence reward mentioned earlier, our reward function includes several terms to regulate the training and preserve high-level semantics. A root tracking reward, denoted as $r^{root}_t=exp(-||\mathbf{s}^{root}_t - \mathbf{\bar{s}}^{root}_t||)$, is introduced to preserve high-level movements by minimizing the deviation between the normalized base trajectory of the human and the robot. Without this term, the resulting correspondence can be arbitrary: e.g., a human forward walking motion can be mapped into a robot's lateral movements.
To prevent unrealistic movements, a torque penalty, $r^{tor}_t=-||\mathbf{a}_t||$, and joint limits penalty, $r^{lim}_t=-\mathbf{1}_{\mathbf{p}^r_t>\mathbf{p}^{lim}}$, are borrowed from \cite{rudin2022learning}. Here, $\mathbf{p}^{lim}$ is the pose limit of the robot. The overall reward is calculated as the weighted sum of all these terms: 
$r_t = w^{cpd}r^{cpd}_t + w^{root}r^{root}_t + w^{tor}r^{tor}_t + w^{lim}r^{lim}_t$.
To optimize the weights, increasing $w^{cpd}$ can improve the correspondence between robot and human motion, but it may negatively affect root tracking performance or increase energy consumption. Increasing $w^{root}$ puts more emphasis on root tracking. Small values for $w^{tor}$ and $w^{lim}$ can result in unnatural motions, while excessively large values can lead to overly conservative motions. In our setting, we use $[w^{cpd}, w^{root}, w^{tor}, w^{lim}] = [1.0, 1.0, 0.0001, 5.0]$.

\textbf{Network Structure.} The policy, critic, H2R-Mapper, and R2H-Mapper are modeled by a fully-connected network consisting of three hidden layers with 512 nodes each. ELU is used as the activation function for the policy and critic, while ReLU is used for the mapper networks.

\section{Experiments}

\begin{figure*}
\vspace{-1em}
\centering
\includegraphics[width=.95\linewidth]{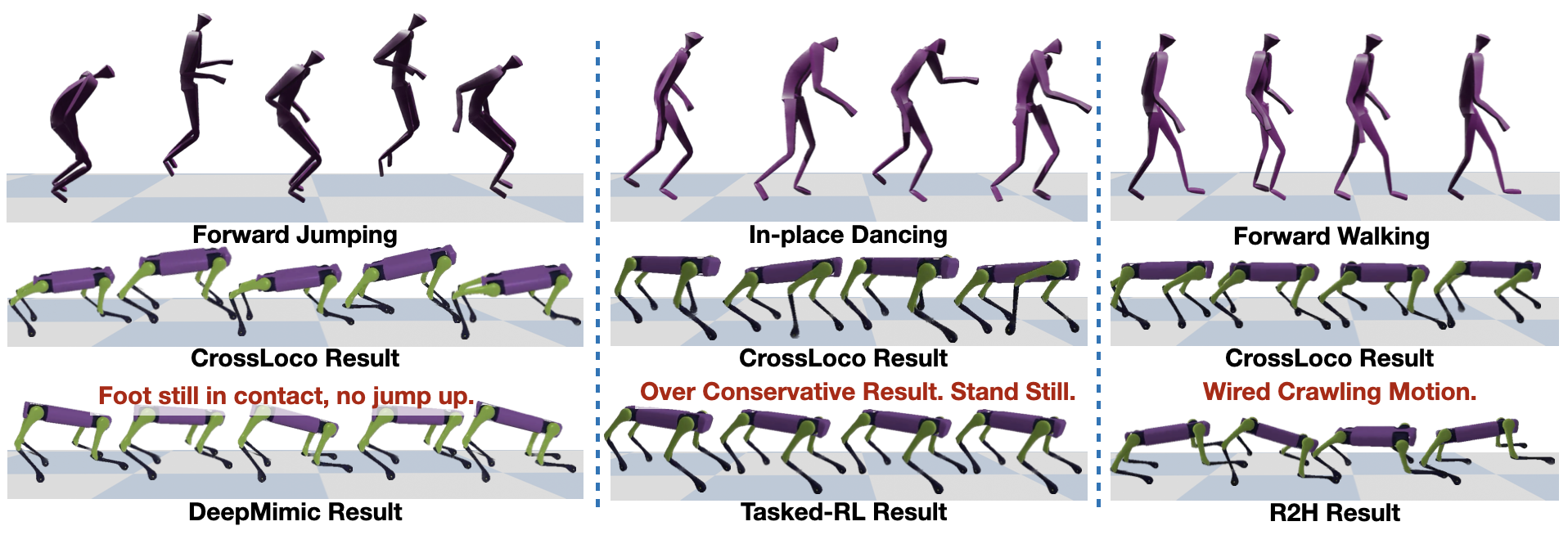}
\caption{Comparison of generated motions. Our method, \textit{CrossLoco}, synthesizes more compelling motions compared to the baselines by preserving both high-level semantics and fine motion details.}
\label{fig:showcase_2}
\end{figure*}
We conduct a series of experiments to investigate three key aspects of the proposed work: firstly, the feasibility of acquiring a human-motion-driven robot controller, which is referred to as the `\textbf{human2robot}' controller; secondly, the comparative performance against alternative baseline approaches; and lastly, the influence of the correspondence reward on the training process.

We evaluate the effectiveness of our approach by transferring a set of human motions to Aliengo quadrupedal robot~\citep{unitree} with 12 joints, which has a significantly different morphology compared to humankind. We take human motion from LaFAN1 dataset~\citep{harvey2020robust}. Our human dataset consists of $50$ human motion trajectories with eight seconds of the average clip length. The dataset contains various types of human movements, including walking, running, hopping, and dancing. 
All experiment environments are conducted using Isaac Gym~\citep{makoviychuk2021isaac}, a high-performance GPU-based physics simulator. During training, $1024$ environments are simulated in parallel on a single NVIDIA GeForce RTX 3080 Ti GPU for a period of about 3 days. 

\subsection{Main Results}
We present the results of our human-driven control experiments in Figure~\ref{fig:showcase_2}. Our method successfully learns a \textit{human2robot} controller that can transfer various human motions to a robot with a different morphology. 
We observe agile motions from the robot, such as running and sharp turning when the human performs fast locomotion. The results also demonstrate that the robot can creatively follow human dancing motions, which is hard to manually design. This showcases the capability of our method to establish automatic correspondence between humans and robots while learning diverse robot skills. All the motions can be best seen in the supplementary video.


In some scenarios, the robot is not able to perfectly mimic the given human motion. For instance, when a human moves backward and makes a sharp 180-degree turn, the robot cannot follow the desired orientation. Additionally, \textit{CrossLoco} may struggle to transfer high-frequency movements, such as hopping. There are two potential reasons for these imperfections. First, the neural network may be incapable of capturing all motions. Second, the robot's morphology may prohibit it from performing certain motions that are easy for humans, such as swift turning.

\subsection{Baseline Comparison}

\begin{table*}
\vspace{-1em}
\begin{tabularx}{0.95\textwidth}{ m{2.7cm} m{1.6cm} m{1.6cm} m{1.6cm} m{1.8cm} m{1.6cm} } 
 \hline 
             & \textbf{ACR} $\uparrow$ & \textbf{DIV} $\uparrow$ & \textbf{RTR} $\uparrow$ & \textbf{CR} $\uparrow$ & \textbf{PR} $\uparrow$  \\ [0.6ex] 

  \hline 
  
 \textbf{CrossLoco (Ours)} & \textbf{0.785} & \textbf{2.853} & \textbf{0.743} & 65.5\% & \textbf{43}\%\\ [0.5ex] 
DeepMimic & 0.558 & 2.231  & 0.579 &  \textbf{67.5}\% & 16\%\\ [0.5ex] 
Task-Only & 0.556 &  2.494 & 0.740 & 30.0\% & 14\%\\ [0.5ex] 
R2H-Only & 0.683 & 2.779 & 0.729 & 42.5\% & 27\%\\ [0.7ex] 
 \hline
\end{tabularx}
\caption{Quantitative Results. Our \textit{CrossLoco} outperforms all the criteria except for being the second-best at correctness with a small margin.} 
\label{tab: baseline_comparison}
\vspace{-0.3cm}
\end{table*}

We further quantitatively compare our method against the following baseline methods:
\begin{itemize}
\item \textbf{Engineered Motion Retarget + DeepMimic~(DeepMimic)}: This baseline contains two stages. Firstly, an expert manually designs a motion retargeting function to translate human motions to robot motion referenced trajectories. Then, the robot is trained to track these reference trajectories using DeepMimic ~ \citep {2018-TOG-deepMimic}. It is important to note that the result of this baseline heavily relies on the quality of the retargeted motions, which requires a significant amount of effort from the expert. In our case, we have designed the retargeted motions by matching the human foot and robot foot with a fixed tripod gait.

\item \textbf{Task-Only}: This baseline is designed to investigate the impact of the proposed correspondence reward on training outcomes. As such, we compare this approach to CrossLoco, where the weight of correspondence is set to zero ($w^x = 0$). Therefore, this policy is trained solely on a root tracking reward and other regularization rewards.

\item \textbf{R2H-Only}: This baseline is designed by removing the robot pose cycle-consistency part from CrossLoco and only keeping the human pose's consistency.  As there is no robot pose consistency, in this baseline, the correspondence reward is defined as  $r^{cpd, r2h}_t = \exp( - ||\ph_{t} -\mu^{r2h}(\pr_{t})  ||^2_2)$.

%

\end{itemize}
Since we assume we have no robot motion dataset, hence, we don't include any GAN-based baseline methods, such as Adversarial Correspondence Embedding~\citep{li2023ace}.

Our objective is to quantitatively assess the effectiveness of establishing correspondence between human and robot motion, as well as the diversity of the robot motion of these methods. In order to achieve this, we utilize the following metrics:
\begin{itemize}
\item \textbf{Averaged Correspondence Reward (ACR)}: This term measures the correspondence between human and robot motions. A higher ACR indicates better correspondence. Even though no correspondence reward is used in each baseline training procedure, we co-train \textit{R2H-Mapper} and \textit{H2R-Mapper} to measure the correspondence reward.
\item \textbf{Diversity (DIV)}: This term has been used for measuring motion diversity in many SOTA works~\citep{shafir2023human, guo2023back}. A higher DIV indicates robots can acquire more skills. From a set of all generated motions from different source human motions, two subsets of the same size $S_d$ are randomly picked.  The diversity of this set of motion is defined as:
$DIV = \frac{1}{S_d}\Sigma^{S_d}_{i=1} ||\Psi(\mathbf{s}^r_i) -\Psi^r(\mathbf{s'}^r_i) ||$. $\Psi(\mathbf{s}^r_i)$ and $\Psi(\mathbf{s'}^r_i)$ are features extracted from robot state. Here, we pick robot root velocity and joint pose as the feature.

\item \textbf{Averaged Root Tracking Reward (RTR)}: This term measures if the learned policy can track the desired root trajectory. The root tracking reward is defined in Section~\ref{sec: implementation_detail}.
\end{itemize}
In addition to these metrics, we conducted a user study with $15$ subjects to evaluate the performance from a subjective perspective.

\begin{itemize}
    \item \textbf{Correct Rate (CR):} We first investigate whether users can identify a correct match between the given human and synthesized robot motions. A user is tasked to find a matched pair from one human animation and four retargeted robot motions. One robot motion is retargeted from the given human motion, while the other three are generated from different inputs. We examined the combination of four human motions and four methods, and then measured the percentage of correct matches.  Ideally, a good transfer should accurately capture the style of the human motion, resulting in easy matching for the user.
    \item \textbf{Preference (PR):} In the second part, we provided users with robot motions generated with different approaches. We asked the users to select the motions that they believed represented a good transfer. We then measured the ratio at which each method was chosen.
\end{itemize}


Our results are summarized in Table ~\ref{tab: baseline_comparison}. The quantitative analysis indicates that \textit{CrossLoco} outperforms the baseline methods in all metrics suggesting that it can effectively learn a controller that can translate different human motions to diverse robot motions while tracking the desired root trajectory. 

Although DeepMimic is designed as a one-to-one mapping between human and robot poses, it achieved a lower correspondence reward than \textit{CrossLoco} (0.558 vs 0.785). This could be attributed to the fact that an engineering mapping function may not be physically feasible for all input human motions, and the learning process may sometimes sacrifice the desired posing tracking for a higher desired root tracking reward. Moreover, since the engineered robot desired motion can sometimes be physically infeasible, the root tracking reward of DeepMimic is also lower than that of CrossLoco (0.579 vs 0.743). Based on the results of the user study, it was found that DeepMimic's motion (with a 16\% PR score) was not preferred by users. However, it achieved the highest CR score (67.5\%). This can be because users can match human and robot motions based on the most obvious frames, even with poor overall motion quality.


As for Task-only, since it is trained only with root tracking and regularization rewards, it produces conservative motions by ignoring human leg motions in some cases, such as different in-place dancing motions.
All human in-place dancing motions are mapped to the robot standing with slight root movements by Task-only baseline. 
However, the correspondence reward in \textit{CrossLoco} triggers robots to learn diverse skills that correspond to different human motions, as evidenced by its superior performance in terms of correspondence reward, diversity term, and user study results compared to Task-only.
We also obverse that for \textit{CrossLoco} achieves slightly higher root tracking reward, indicating \textit{CrossLoco}'s great capability.

The results show that CrossLoco achieves a higher correspondence reward and more diverse motion compared to R2H-Only. Additionally, users found the results of CrossLoco to be more distinguishable. This could be attributed to the effective regularization provided by the cycling of human back to robot, resulting in more distinguishable outcomes. 

\section{Applications}

Many research and work have explored the field of human motion synthesis using various input sources, including text \citep{bahl2022human}, music \cite{tseng2023edge}, and user inputs \citep{holden2020learned}. Our learned human2robot controller can be seamlessly integrated with these modules by using human motion as the interface for new applications. In this section, we present two examples: \textbf{Language2Robot} and \textbf{Interactive Robot Control}.

\textbf{Language2Robot.} Our approach involves utilizing the text2human module in combination with our human2robot controller. This allows for the generation of robot movements from language by first producing human motion using the text2human module, which is then transferred to the robot using the human2robot controller. Our method differs from recent language to quadrupedal robot motion work~\citep{tang2023saytap} in that it does not require an engineering-intensive interface, such as foot contact patterns, which could limit the range of possible generated robot motions.

In our implementation, we utilize the Human Motion Diffusion Model (MDM)~\citep{bahl2022human} as our text-to-human motion translator. MDM is a diffusion model-based lightweight model that achieves state-of-the-art results on leading benchmarks. However, MDM uses AMASS~\citep{mahmood2019amass} human model which is different from the LaFAN1~\citep{harvey2020robust} model we used for training the policy. Therefore, we retarget the outputs of the MDM to the skeleton model we use. 

\begin{figure*}[t]
 \vspace{-2.5em}
 \centering
    \begin{minipage}[b]{.995\textwidth}
        \centering
        \includegraphics[width=.9\linewidth]{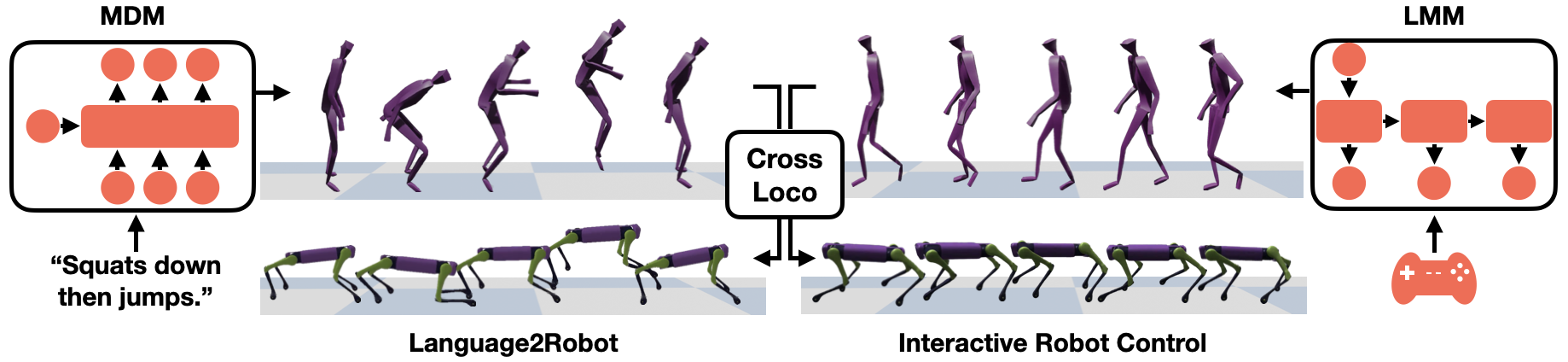}
        \label{fig:MDM_exp}
    \end{minipage}

\caption{Illustration of two applications: \textbf{Language2Robot} and \textbf{Interactive Robot Control}.}
\label{fig: Application results}
 \vspace{-1em}
\end{figure*}

Our study involves testing the generation of robot motion based on different input messages. The results of our study are presented in Figure~\ref{fig: Application results}. We show that this framework can generate robot motion according to instructions. For instance, the message ``strides swiftly in a straight'' results in a fast walking straight robot motion, while ``squats down then jumps'' triggers a robot squats and jumps motion.

\textbf{Interactive Robot Control.} Interactive control of robots in response to changing conditions or unexpected obstacles is a significant challenge in robotics, which involves careful controller design and motion planning. Recent advances in character animation enable users to interactively control human characters using joysticks, automatically adapting their motion styles to the surrounding environment, like crouching in confined spaces or leaping over obstacles. Our key idea for the second application, \textbf{Interactive Robot Control}, is to leverage the existing human animation techniques for robot control. Instead of retraining a large-scale model from scratch, we simply translate the output of the existing character controller to the robot's operational space using the proposed method.

Our implementation of this framework utilizes Learned Motion Matching (LMM)~\citep{holden2020learned}, which is a scalable neural network-based framework for interactive motion synthesis. We combine LMM with our learned controller. During interactive robot control, LMM takes input user commands for human motion generation, and our controller converts the generated human motion to robot control commands.

We evaluated our implementation by controlling the robot using a joystick. Figure~\ref{fig: Application results} presents the results of using LMM2robot. The experiment demonstrated that the robot can actively adjust its motion based on the user's commands. These results provide evidence of the effectiveness of our learned controller for interactive robot control.

\section{Discussion}
We introduce CrossLoco, an unsupervised reinforcement learning framework designed to enable robot locomotion control driven by human motion. This framework incorporates a cycle-consistency-based reward function, which facilitates the discovery of robot skills and the establishment of correspondence between human and robot motion. Our experimental results demonstrate the effectiveness of our approach in transferring a wide range of human motions to control a robot that has a different morphology.

Our next steps involve exploring two directions. Firstly, we aim to extend our framework beyond locomotion control to more complex scenarios, including long-horizon human demonstrations that involve long-distance locomotion and tool manipulation on a legged-manipulation robot, such as a quadrupedal robot with an arm mounted on its body. Secondly, we are interested in implementing our method on real-world robots for practical applications.

\bibliography{iclr2024_conference}

\begin{thebibliography}{57}
\providecommand{\natexlab}[1]{#1}
\providecommand{\url}[1]{\texttt{#1}}
\expandafter\ifx\csname urlstyle\endcsname\relax
  \providecommand{\doi}[1]{doi: #1}\else
  \providecommand{\doi}{doi: \begingroup \urlstyle{rm}\Url}\fi

\bibitem[Aberman et~al.(2020)Aberman, Li, Lischinski, Sorkine-Hornung,
  Cohen-Or, and Chen]{aberman2020skeleton}
Kfir Aberman, Peizhuo Li, Dani Lischinski, Olga Sorkine-Hornung, Daniel
  Cohen-Or, and Baoquan Chen.
\newblock Skeleton-aware networks for deep motion retargeting.
\newblock \emph{ACM Transactions on Graphics (TOG)}, 39\penalty0 (4):\penalty0
  62--1, 2020.

\bibitem[Bahl et~al.(2022)Bahl, Gupta, and Pathak]{bahl2022human}
Shikhar Bahl, Abhinav Gupta, and Deepak Pathak.
\newblock Human-to-robot imitation in the wild.
\newblock 2022.

\bibitem[Bansal et~al.(2018)Bansal, Ma, Ramanan, and Sheikh]{bansal2018recycle}
Aayush Bansal, Shugao Ma, Deva Ramanan, and Yaser Sheikh.
\newblock Recycle-gan: Unsupervised video retargeting.
\newblock In \emph{Proceedings of the European conference on computer vision
  (ECCV)}, pp.\  119--135, 2018.

\bibitem[Bergamin et~al.(2019)Bergamin, Clavet, Holden, and
  Forbes]{bergamin2019drecon}
Kevin Bergamin, Simon Clavet, Daniel Holden, and James~Richard Forbes.
\newblock Drecon: data-driven responsive control of physics-based characters.
\newblock \emph{ACM Transactions On Graphics (TOG)}, 38\penalty0 (6):\penalty0
  1--11, 2019.

\bibitem[Bousmalis et~al.(2018)Bousmalis, Irpan, Wohlhart, Bai, Kelcey,
  Kalakrishnan, Downs, Ibarz, Pastor, Konolige, et~al.]{bousmalis2018using}
Konstantinos Bousmalis, Alex Irpan, Paul Wohlhart, Yunfei Bai, Matthew Kelcey,
  Mrinal Kalakrishnan, Laura Downs, Julian Ibarz, Peter Pastor, Kurt Konolige,
  et~al.
\newblock Using simulation and domain adaptation to improve efficiency of deep
  robotic grasping.
\newblock In \emph{2018 IEEE international conference on robotics and
  automation (ICRA)}, pp.\  4243--4250. IEEE, 2018.

\bibitem[Choi et~al.(2020)Choi, Pan, and Kim]{choi2020nonparametric}
Sungjoon Choi, Matt Pan, and Joohyung Kim.
\newblock Nonparametric motion retargeting for humanoid robots on shared latent
  space.
\newblock In \emph{16th Robotics: Science and Systems, RSS 2020}. MIT Press
  Journals, 2020.

\bibitem[del Toro(2013)]{Pacific_Rim}
Guillermo del Toro.
\newblock Pacific rim, 2013.

\bibitem[Delhaisse et~al.(2017)Delhaisse, Esteban, Rozo, and
  Caldwell]{delhaisse2017transfer}
Brian Delhaisse, Domingo Esteban, Leonel Rozo, and Darwin Caldwell.
\newblock Transfer learning of shared latent spaces between robots with similar
  kinematic structure.
\newblock In \emph{2017 International Joint Conference on Neural Networks
  (IJCNN)}, pp.\  4142--4149. IEEE, 2017.

\bibitem[Eysenbach et~al.(2018)Eysenbach, Gupta, Ibarz, and
  Levine]{eysenbach2018diversity}
Benjamin Eysenbach, Abhishek Gupta, Julian Ibarz, and Sergey Levine.
\newblock Diversity is all you need: Learning skills without a reward function.
\newblock \emph{arXiv preprint arXiv:1802.06070}, 2018.

\bibitem[Gleicher(1998)]{gleicher1998retargetting}
Michael Gleicher.
\newblock Retargetting motion to new characters.
\newblock In \emph{Proceedings of the 25th annual conference on Computer
  graphics and interactive techniques}, pp.\  33--42, 1998.

\bibitem[Goodfellow et~al.(2014)Goodfellow, Pouget-Abadie, Mirza, Xu,
  Warde-Farley, Ozair, Courville, and Bengio]{goodfellow2014generative}
Ian Goodfellow, Jean Pouget-Abadie, Mehdi Mirza, Bing Xu, David Warde-Farley,
  Sherjil Ozair, Aaron Courville, and Yoshua Bengio.
\newblock Generative adversarial nets.
\newblock \emph{Advances in neural information processing systems}, 27, 2014.

\bibitem[Grandia et~al.(2023)Grandia, Farshidian, Knoop, Schumacher, Hutter,
  and B{\"a}cher]{grandia2023doc}
Ruben Grandia, Farbod Farshidian, Espen Knoop, Christian Schumacher, Marco
  Hutter, and Moritz B{\"a}cher.
\newblock Doc: Differentiable optimal control for retargeting motions onto
  legged robots.
\newblock \emph{ACM Transactions on Graphics}, 42\penalty0 (4), 2023.

\bibitem[Gregor et~al.(2016)Gregor, Rezende, and
  Wierstra]{gregor2016variational}
Karol Gregor, Danilo~Jimenez Rezende, and Daan Wierstra.
\newblock Variational intrinsic control.
\newblock \emph{arXiv preprint arXiv:1611.07507}, 2016.

\bibitem[Guo et~al.(2023)Guo, Du, Shen, Lepetit, Alameda-Pineda, and
  Moreno-Noguer]{guo2023back}
Wen Guo, Yuming Du, Xi~Shen, Vincent Lepetit, Xavier Alameda-Pineda, and
  Francesc Moreno-Noguer.
\newblock Back to mlp: A simple baseline for human motion prediction.
\newblock In \emph{Proceedings of the IEEE/CVF Winter Conference on
  Applications of Computer Vision}, pp.\  4809--4819, 2023.

\bibitem[Haarnoja et~al.(2018)Haarnoja, Ha, Zhou, Tan, Tucker, and
  Levine]{haarnoja2018learning}
Tuomas Haarnoja, Sehoon Ha, Aurick Zhou, Jie Tan, George Tucker, and Sergey
  Levine.
\newblock Learning to walk via deep reinforcement learning.
\newblock \emph{arXiv preprint arXiv:1812.11103}, 2018.

\bibitem[Harvey et~al.(2020)Harvey, Yurick, Nowrouzezahrai, and
  Pal]{harvey2020robust}
Félix~G. Harvey, Mike Yurick, Derek Nowrouzezahrai, and Christopher Pal.
\newblock Robust motion in-betweening.
\newblock 39\penalty0 (4), 2020.

\bibitem[Hejna et~al.(2020)Hejna, Pinto, and Abbeel]{hejna2020hierarchically}
Donald Hejna, Lerrel Pinto, and Pieter Abbeel.
\newblock Hierarchically decoupled imitation for morphological transfer.
\newblock In \emph{International Conference on Machine Learning}, pp.\
  4159--4171. PMLR, 2020.

\bibitem[Ho \& Ermon(2016)Ho and Ermon]{ho2016generative}
Jonathan Ho and Stefano Ermon.
\newblock Generative adversarial imitation learning.
\newblock \emph{Advances in neural information processing systems}, 29, 2016.

\bibitem[Hoffman et~al.(2018)Hoffman, Tzeng, Park, Zhu, Isola, Saenko, Efros,
  and Darrell]{hoffman2018cycada}
Judy Hoffman, Eric Tzeng, Taesung Park, Jun-Yan Zhu, Phillip Isola, Kate
  Saenko, Alexei Efros, and Trevor Darrell.
\newblock Cycada: Cycle-consistent adversarial domain adaptation.
\newblock In \emph{International conference on machine learning}, pp.\
  1989--1998. Pmlr, 2018.

\bibitem[Holden et~al.(2020)Holden, Kanoun, Perepichka, and
  Popa]{holden2020learned}
Daniel Holden, Oussama Kanoun, Maksym Perepichka, and Tiberiu Popa.
\newblock Learned motion matching.
\newblock \emph{ACM Transactions on Graphics (TOG)}, 39\penalty0 (4):\penalty0
  53--1, 2020.

\bibitem[Iuchi(1989)]{Madö_King_Granzört}
Shūji Iuchi.
\newblock Madö king granzört, 1989.

\bibitem[James et~al.(2019)James, Wohlhart, Kalakrishnan, Kalashnikov, Irpan,
  Ibarz, Levine, Hadsell, and Bousmalis]{james2019sim}
Stephen James, Paul Wohlhart, Mrinal Kalakrishnan, Dmitry Kalashnikov, Alex
  Irpan, Julian Ibarz, Sergey Levine, Raia Hadsell, and Konstantinos Bousmalis.
\newblock Sim-to-real via sim-to-sim: Data-efficient robotic grasping via
  randomized-to-canonical adaptation networks.
\newblock In \emph{Proceedings of the IEEE/CVF Conference on Computer Vision
  and Pattern Recognition}, pp.\  12627--12637, 2019.

\bibitem[Jang et~al.(2018)Jang, Kwon, Yu, Kim, and Kim]{jang2018variational}
Hanyoung Jang, Byungjun Kwon, Moonwon Yu, Seong~Uk Kim, and Jongmin Kim.
\newblock A variational u-net for motion retargeting.
\newblock In \emph{SIGGRAPH Asia 2018 Posters}, pp.\  1--2. 2018.

\bibitem[Kim et~al.(2020)Kim, Gu, Song, Zhao, and Ermon]{kim2020domain}
Kuno Kim, Yihong Gu, Jiaming Song, Shengjia Zhao, and Stefano Ermon.
\newblock Domain adaptive imitation learning.
\newblock In \emph{International Conference on Machine Learning}, pp.\
  5286--5295. PMLR, 2020.

\bibitem[Li et~al.(2019)Li, Geyer, Atkeson, and Rai]{li2019using}
Tianyu Li, Hartmut Geyer, Christopher~G Atkeson, and Akshara Rai.
\newblock Using deep reinforcement learning to learn high-level policies on the
  atrias biped.
\newblock In \emph{2019 International Conference on Robotics and Automation
  (ICRA)}, pp.\  263--269. IEEE, 2019.

\bibitem[Li et~al.(2023{\natexlab{a}})Li, Won, Cho, Ha, and
  Rai]{li2023fastmimic}
Tianyu Li, Jungdam Won, Jeongwoo Cho, Sehoon Ha, and Akshara Rai.
\newblock Fastmimic: Model-based motion imitation for agile, diverse and
  generalizable quadrupedal locomotion.
\newblock \emph{Robotics}, 12\penalty0 (3):\penalty0 90, 2023{\natexlab{a}}.

\bibitem[Li et~al.(2023{\natexlab{b}})Li, Won, Clegg, Kim, Rai, and
  Ha]{li2023ace}
Tianyu Li, Jungdam Won, Alexander Clegg, Jeonghwan Kim, Akshara Rai, and Sehoon
  Ha.
\newblock Ace: Adversarial correspondence embedding for cross morphology motion
  retargeting from human to nonhuman characters.
\newblock \emph{SIGGRAPH Asia}, 2023{\natexlab{b}}.

\bibitem[Ling et~al.(2020)Ling, Zinno, Cheng, and Van
  De~Panne]{ling2020character}
Hung~Yu Ling, Fabio Zinno, George Cheng, and Michiel Van De~Panne.
\newblock Character controllers using motion vaes.
\newblock \emph{ACM Transactions on Graphics (TOG)}, 39\penalty0 (4):\penalty0
  40--1, 2020.

\bibitem[Liu et~al.(2017)Liu, Breuel, and Kautz]{liu2017unsupervised}
Ming-Yu Liu, Thomas Breuel, and Jan Kautz.
\newblock Unsupervised image-to-image translation networks.
\newblock \emph{Advances in neural information processing systems}, 30, 2017.

\bibitem[Mahmood et~al.(2019)Mahmood, Ghorbani, Troje, Pons-Moll, and
  Black]{mahmood2019amass}
Naureen Mahmood, Nima Ghorbani, Nikolaus~F Troje, Gerard Pons-Moll, and
  Michael~J Black.
\newblock Amass: Archive of motion capture as surface shapes.
\newblock In \emph{Proceedings of the IEEE/CVF international conference on
  computer vision}, pp.\  5442--5451, 2019.

\bibitem[Makoviychuk et~al.(2021)Makoviychuk, Wawrzyniak, Guo, Lu, Storey,
  Macklin, Hoeller, Rudin, Allshire, Handa, et~al.]{makoviychuk2021isaac}
Viktor Makoviychuk, Lukasz Wawrzyniak, Yunrong Guo, Michelle Lu, Kier Storey,
  Miles Macklin, David Hoeller, Nikita Rudin, Arthur Allshire, Ankur Handa,
  et~al.
\newblock Isaac gym: High performance gpu-based physics simulation for robot
  learning.
\newblock \emph{arXiv preprint arXiv:2108.10470}, 2021.

\bibitem[Peng et~al.(2018{\natexlab{a}})Peng, Abbeel, Levine, and van~de
  Panne]{2018-TOG-deepMimic}
Xue~Bin Peng, Pieter Abbeel, Sergey Levine, and Michiel van~de Panne.
\newblock Deepmimic: Example-guided deep reinforcement learning of
  physics-based character skills.
\newblock \emph{ACM Trans. Graph.}, 37\penalty0 (4):\penalty0 143:1--143:14,
  July 2018{\natexlab{a}}.
\newblock ISSN 0730-0301.
\newblock \doi{10.1145/3197517.3201311}.
\newblock URL \url{http://doi.acm.org/10.1145/3197517.3201311}.

\bibitem[Peng et~al.(2018{\natexlab{b}})Peng, Kanazawa, Malik, Abbeel, and
  Levine]{2018-TOG-SFV}
Xue~Bin Peng, Angjoo Kanazawa, Jitendra Malik, Pieter Abbeel, and Sergey
  Levine.
\newblock Sfv: Reinforcement learning of physical skills from videos.
\newblock \emph{ACM Trans. Graph.}, 37\penalty0 (6), November
  2018{\natexlab{b}}.

\bibitem[Peng et~al.(2020)Peng, Coumans, Zhang, Lee, Tan, and
  Levine]{RoboImitationPeng20}
Xue~Bin Peng, Erwin Coumans, Tingnan Zhang, Tsang-Wei~Edward Lee, Jie Tan, and
  Sergey Levine.
\newblock Learning agile robotic locomotion skills by imitating animals.
\newblock In \emph{Robotics: Science and Systems}, 07 2020.
\newblock \doi{10.15607/RSS.2020.XVI.064}.

\bibitem[Peng et~al.(2022)Peng, Guo, Halper, Levine, and Fidler]{2022-TOG-ASE}
Xue~Bin Peng, Yunrong Guo, Lina Halper, Sergey Levine, and Sanja Fidler.
\newblock Ase: Large-scale reusable adversarial skill embeddings for physically
  simulated characters.
\newblock \emph{ACM Trans. Graph.}, 41\penalty0 (4), July 2022.

\bibitem[Rao et~al.(2020)Rao, Harris, Irpan, Levine, Ibarz, and
  Khansari]{rao2020rl}
Kanishka Rao, Chris Harris, Alex Irpan, Sergey Levine, Julian Ibarz, and Mohi
  Khansari.
\newblock Rl-cyclegan: Reinforcement learning aware simulation-to-real.
\newblock In \emph{Proceedings of the IEEE/CVF Conference on Computer Vision
  and Pattern Recognition}, pp.\  11157--11166, 2020.

\bibitem[Rhodin et~al.(2014)Rhodin, Tompkin, In~Kim, Varanasi, Seidel, and
  Theobalt]{rhodin2014interactive}
Helge Rhodin, James Tompkin, Kwang In~Kim, Kiran Varanasi, Hans-Peter Seidel,
  and Christian Theobalt.
\newblock Interactive motion mapping for real-time character control.
\newblock In \emph{Computer Graphics Forum}, volume~33, pp.\  273--282. Wiley
  Online Library, 2014.

\bibitem[Rudin et~al.(2022)Rudin, Hoeller, Reist, and
  Hutter]{rudin2022learning}
Nikita Rudin, David Hoeller, Philipp Reist, and Marco Hutter.
\newblock Learning to walk in minutes using massively parallel deep
  reinforcement learning.
\newblock In \emph{Conference on Robot Learning}, pp.\  91--100. PMLR, 2022.

\bibitem[Schulman et~al.(2017)Schulman, Wolski, Dhariwal, Radford, and
  Klimov]{schulman2017proximal}
John Schulman, Filip Wolski, Prafulla Dhariwal, Alec Radford, and Oleg Klimov.
\newblock Proximal policy optimization algorithms.
\newblock \emph{arXiv preprint arXiv:1707.06347}, 2017.

\bibitem[Sermanet et~al.(2018)Sermanet, Lynch, Chebotar, Hsu, Jang, Schaal,
  Levine, and Brain]{sermanet2018time}
Pierre Sermanet, Corey Lynch, Yevgen Chebotar, Jasmine Hsu, Eric Jang, Stefan
  Schaal, Sergey Levine, and Google Brain.
\newblock Time-contrastive networks: Self-supervised learning from video.
\newblock In \emph{2018 IEEE international conference on robotics and
  automation (ICRA)}, pp.\  1134--1141. IEEE, 2018.

\bibitem[Shafir et~al.(2023)Shafir, Tevet, Kapon, and Bermano]{shafir2023human}
Yonatan Shafir, Guy Tevet, Roy Kapon, and Amit~H Bermano.
\newblock Human motion diffusion as a generative prior.
\newblock \emph{arXiv preprint arXiv:2303.01418}, 2023.

\bibitem[Shankar et~al.(2022)Shankar, Lin, Rajeswaran, Kumar, Anderson, and
  Oh]{shankar2022translating}
Tanmay Shankar, Yixin Lin, Aravind Rajeswaran, Vikash Kumar, Stuart Anderson,
  and Jean Oh.
\newblock Translating robot skills: Learning unsupervised skill correspondences
  across robots.
\newblock In \emph{International Conference on Machine Learning}, pp.\
  19626--19644. PMLR, 2022.

\bibitem[Sharma et~al.(2019)Sharma, Pathak, and Gupta]{sharma2019third}
Pratyusha Sharma, Deepak Pathak, and Abhinav Gupta.
\newblock Third-person visual imitation learning via decoupled hierarchical
  controller.
\newblock \emph{Advances in Neural Information Processing Systems}, 32, 2019.

\bibitem[Smith et~al.(2019)Smith, Dhawan, Zhang, Abbeel, and
  Levine]{smith2019avid}
Laura Smith, Nikita Dhawan, Marvin Zhang, Pieter Abbeel, and Sergey Levine.
\newblock Avid: Learning multi-stage tasks via pixel-level translation of human
  videos.
\newblock \emph{arXiv preprint arXiv:1912.04443}, 2019.

\bibitem[Spielberg(2018)]{Ready_Player_One}
Steven Spielberg.
\newblock Ready player one, 2018.

\bibitem[Stein \& Roy(2018)Stein and Roy]{stein2018genesis}
Gregory~J Stein and Nicholas Roy.
\newblock Genesis-rt: Generating synthetic images for training secondary
  real-world tasks.
\newblock In \emph{2018 IEEE International Conference on Robotics and
  Automation (ICRA)}, pp.\  7151--7158. IEEE, 2018.

\bibitem[Tak \& Ko(2005)Tak and Ko]{tak2005physically}
Seyoon Tak and Hyeong-Seok Ko.
\newblock A physically-based motion retargeting filter.
\newblock \emph{ACM Transactions on Graphics (TOG)}, 24\penalty0 (1):\penalty0
  98--117, 2005.

\bibitem[Tan et~al.(2018)Tan, Zhang, Coumans, Iscen, Bai, Hafner, Bohez, and
  Vanhoucke]{tan2018sim}
Jie Tan, Tingnan Zhang, Erwin Coumans, Atil Iscen, Yunfei Bai, Danijar Hafner,
  Steven Bohez, and Vincent Vanhoucke.
\newblock Sim-to-real: Learning agile locomotion for quadruped robots.
\newblock \emph{arXiv preprint arXiv:1804.10332}, 2018.

\bibitem[Tang et~al.(2023)Tang, Yu, Tan, Zen, Faust, and
  Harada]{tang2023saytap}
Yujin Tang, Wenhao Yu, Jie Tan, Heiga Zen, Aleksandra Faust, and Tatsuya
  Harada.
\newblock Saytap: Language to quadrupedal locomotion.
\newblock \emph{arXiv preprint arXiv:2306.07580}, 2023.

\bibitem[Tseng et~al.(2023)Tseng, Castellon, and Liu]{tseng2023edge}
Jonathan Tseng, Rodrigo Castellon, and Karen Liu.
\newblock Edge: Editable dance generation from music.
\newblock In \emph{Proceedings of the IEEE/CVF Conference on Computer Vision
  and Pattern Recognition}, pp.\  448--458, 2023.

\bibitem[unitree(2023)]{unitree}
unitree.
\newblock Unitree robotics, 2023.

\bibitem[Villegas et~al.(2018)Villegas, Yang, Ceylan, and
  Lee]{villegas2018neural}
Ruben Villegas, Jimei Yang, Duygu Ceylan, and Honglak Lee.
\newblock Neural kinematic networks for unsupervised motion retargetting.
\newblock In \emph{Proceedings of the IEEE conference on computer vision and
  pattern recognition}, pp.\  8639--8648, 2018.

\bibitem[Won \& Lee(2019)Won and Lee]{won2019learning}
Jungdam Won and Jehee Lee.
\newblock Learning body shape variation in physics-based characters.
\newblock \emph{ACM Transactions on Graphics (TOG)}, 38\penalty0 (6):\penalty0
  1--12, 2019.

\bibitem[Xie et~al.(2018)Xie, Berseth, Clary, Hurst, and van~de
  Panne]{xie2018feedback}
Zhaoming Xie, Glen Berseth, Patrick Clary, Jonathan Hurst, and Michiel van~de
  Panne.
\newblock Feedback control for cassie with deep reinforcement learning.
\newblock In \emph{2018 IEEE/RSJ International Conference on Intelligent Robots
  and Systems (IROS)}, pp.\  1241--1246. IEEE, 2018.

\bibitem[Zhang et~al.(2020)Zhang, Xiao, Efros, Pinto, and
  Wang]{zhang2020learning}
Qiang Zhang, Tete Xiao, Alexei~A Efros, Lerrel Pinto, and Xiaolong Wang.
\newblock Learning cross-domain correspondence for control with dynamics
  cycle-consistency.
\newblock \emph{arXiv preprint arXiv:2012.09811}, 2020.

\bibitem[Zhou et~al.(2016)Zhou, Krahenbuhl, Aubry, Huang, and
  Efros]{zhou2016learning}
Tinghui Zhou, Philipp Krahenbuhl, Mathieu Aubry, Qixing Huang, and Alexei~A
  Efros.
\newblock Learning dense correspondence via 3d-guided cycle consistency.
\newblock In \emph{Proceedings of the IEEE conference on computer vision and
  pattern recognition}, pp.\  117--126, 2016.

\bibitem[Zhu et~al.(2017)Zhu, Park, Isola, and Efros]{zhu2017unpaired}
Jun-Yan Zhu, Taesung Park, Phillip Isola, and Alexei~A Efros.
\newblock Unpaired image-to-image translation using cycle-consistent
  adversarial networks.
\newblock In \emph{Proceedings of the IEEE international conference on computer
  vision}, pp.\  2223--2232, 2017.

\end{thebibliography}
\bibliographystyle{iclr2024_conference}

\end{document}